# An Artificial Intelligence-Based System for Nutrient Intake Assessment of Hospitalised Patients*

Ya Lu, Thomai Stathopoulou, Maria F. Vasiloglou, Stergios Christodoulidis, Beat Blum, Thomas Walser, Vinzenz Meier, Zeno Stanga, Stavroula G. Mougiakakou, *Member, IEEE*

*Abstract*—Regular nutrient intake monitoring in hospitalised patients plays a critical role in reducing the risk of disease-related malnutrition (DRM). Although several methods to estimate nutrient intake have been developed, there is still a clear demand for a more reliable and fully automated technique, as this could improve the data accuracy and reduce both the participant burden and the health costs. In this paper, we propose a novel system based on artificial intelligence to accurately estimate nutrient intake, by simply processing RGB depth image pairs captured before and after a meal consumption. For the development and evaluation of the system, a dedicated and new database of images and recipes of 322 meals was assembled, coupled to data annotation using innovative strategies. With this database, a system was developed that employed a novel multi-task neural network and an algorithm for 3D surface construction. This allowed sequential semantic food segmentation and estimation of the volume of the consumed food, and permitted fully automatic estimation of nutrient intake for each food type with a 15% estimation error.

## I. INTRODUCTION

DRM (disease-related malnutrition) is a serious condition, which is associated with an increased risk of hospital infections, higher mortality, morbidity, prolonged length of hospital stay and extra healthcare expenses [1]. Investigations in seven Swiss hospitals have shown that 20% of patients were severely undernourished or "at risk" of malnutrition [2]. Thus, maintaining a good nutritional status is of vital importance for both hospitalised patients and social medical systems.

The risk of hospitalised malnutrition is mainly due to poor recognition and monitoring of nutritional intake [3]. To address these problems, it is essential that hospitalised patients' food intake is regularly evaluated. Suggested approaches include weighing plate waste [4], visual estimations [5], digital photography [6] and combined use of digital camera and weighing scale [7]. Nevertheless, functions to identify and estimate portion size are not yet advanced enough to allow fully automated data analysis. Therefore, there is still an unambiguous need for a reliable and simple solution to estimate nutrient intake.

In this paper, we propose an AI-based, fully automatic monitoring system for nutrient intake by hospitalised patients, by analysing the RGB-D image pairs captured before and after the meal. A new database of food images with nutrient recipes has been built up in collaboration with the central kitchen of the Bern University Hospital, in order to develop and evaluate the system. In such a database, pixel-level semantic segmentation maps for both food and plates are annotated, in order to make full use of the information and thus to maximise performance. Based on this database, a dedicated system was designed to monitor nutrient intake, involving four stages: 1) semantic food segmentation using a novel multi-task neural network; 2) segmentation map refinement utilising statistical modelling [8] and context information; 3) estimation of volume of food consumed, using an algorithm for construction of the 3D surface [9]; and 4) calculation of nutrient intake from the consumed volume, food type and the corresponding menu and nutrient recipes, with errors as low as only ca. 15%.

## II. DATA SET

For the design, development and evaluation of the various components of the proposed system, a dedicated and novel database named "Inselspital Nutrient Intake Monitoring Database" (INIMD) has been compiled, that contains RGB-D image pairs of 322 real world meals for patients, with 521 food categories in total. All the meals - as well as the associated nutrient information and recipes - were provided by the central kitchen of the university hospital of Bern.

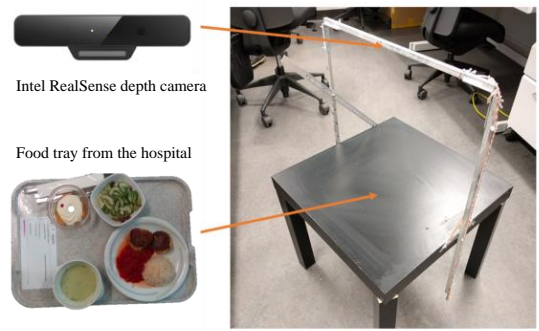

Fig. 1. Data capture prototype.

*Research supported by SV Stiftung Foundation.

Ya Lu, Thomai Stathopoulou, Maria F. Vasiloglou, Stergios Christodoulidis and Stavroula Mougiakakou are with the ARTORG Center for Biomedical Engineering Research, University of Bern. Murtenstrasse 50, CH-3008 Bern, Switzerland (phone: +41-31-632-7592; e-mail: {ya.lu, thomai.stathopoulou, maria.vasiloglou, stergios.christodoulidis, stavroula.mougiakakou}@artorg.unibe.ch).

Beat Blum, Thomas Walser and Vinzenz Meier are with the Department of Catering and the Central Kitchen of the University Hospital Bern. Bern, Switzerland (e-mail: {beat.blum, thomas.walser, vinzenz.meier}@insel.ch).

Zeno Stanga is with the Department of Diabetes, Endocrinology, Clinical Nutrition and Metabolism (UDEM), Inselspital, Bern University Hospital, University of Bern, Freiburgstrasse 15, CH-3010 Bern, Switzerland (e-mail: {Zeno.Stanga}@insel.ch).

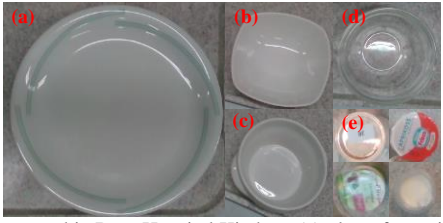

Fig. 2 Plates used in Bern Hospital Kitchen; (a) plates for main course, vegetable, side dish; (b): salad bowl; (c) soup bowl; (d): dessert bowl; (e) other packaged containers.

## A. Data capture

The images were captured using the Intel RealSense RGB-D sensor that was attached on a metal frame fixed on a table, as shown in Fig. 1. The distance between the camera and the meal placed on the table is around 40 cm, and the resolution of each acquired RGB-D image pair is 640×480.

To ensure that the database is highly diverse, the acquisition of the image pairs conducted in two stages. At first, 153 RGB-D image pairs were captured from 153 food trays only before each meal; then another 507 RGB-D image pairs were captured from the remaining 169 food trays, before, in the middle of and after each meal. While capturing all these 660 RGB-D image pairs, the weight of each plate inside the food tray was recorded and will be hereafter used as the nutrient intake ground truth.

## B. Data Annotation

In order to generate the ground truth segmentation for each tray image, the pixel-level food semantic segmentation map was annotated, using an in-house developed image annotation tool. According to the menus provided by the central kitchen, we annotated the food items on the tray into seven (7) hyper categories: soup, main course, sauce, vegetable, side dish, salad and dessert.

In addition, we segmented the different plate types, classifying the plates into five (5) categories: main plate, salad bowl, soup bowl, dessert bowl, and other packaged containers, as exemplified in Fig. 2. It should be emphasised that the use of such plate segmentation map can greatly improve the performance of the proposed system in three aspects: 1) allowing use of a novel multi-task learning approach to improve the accuracy of food segmentation; 2) providing the context information to further refine the food semantic segmentation map in the post-processing stage; 3) enabling estimation of the depth of plate surface in order to correctly calculate food volume. All these benefits will be demonstrated in Section IV.

## III. METHOD

In Fig. 3 the overall architecture of the proposed methodology is illustrated. In the following sections the four main stages of the proposed system are presented.

### A. Image Segmentation Network

A Multi-Task Fully Convolutional Network (MTFCNet) is proposed as the image segmentation module, which takes each RGB-D image pair as input, and simultaneously outputs the types of food and plate in two respective segmentation maps.

The network we designed is mainly based on an encoder-decoder architecture with skip connections. Table I shows the detailed configurations of the network, in which the convolutional layer is denoted as $C^i k/f/s$, where $i$ is the layer index, k is the kernel size, f is the number of filters and s indicates stride. Similarly, each deconvolutional layer is denoted as $DC^i k/f/s$, while the skip connected convolutional layer is represented by $SCC^{ei} k/f/s$, where $ei$ is the corresponding layer index in the Encoder component. Note that for both components, all the residual convolutional/deconvolutional layers are followed by batch normalisation layers and ReLU activation – with the exception of the last layer, that is connected with the softmax activation layer to predict results.

TABLE I. SEGMENTATION NETWORK ARCHITECTURE

| Model Component | Structure |
|---|---|
| Encoder | $C^1 3/16/2 - C^2 3/32/2 - C^3 3/64/2 - C^4 3/128/2 - C^5 3/256/2 - C^6 3/512/1$ |
| Decode Food | $DC^1 3/512/2 - SCC^4 3/512/1 - DC^2 3/256/2 - SCC^3 3/256/1 - DC^3 3/128/2 - SCC^2 3/128/1 - DC^4 3/64/2 - SCC^1 3/64/1 - DC^5 3/32/2 - C^7 1/8/1 \rightarrow$ food seg. map |
| Decode Plate | $DC^6 3/512/2 - SCC^4 3/512/1 - DC^7 3/256/2 - SCC^3 3/256/1 - DC^8 3/128/2 - SCC^2 3/128/1 - DC^9 3/64/2 - SCC^1 3/64/1 - DC^{10} 3/32/2 - C^8 3/6/1 \rightarrow$ plate type map |

### B. Food Segments Refinement

Although this network is able to output an initial segmentation map of hyper food category, errors still exist in practical scenarios, due to variations in lighting, noisy background or inter-class similarity. Two types of post-processing methods, based on statistical modelling and context information, are therefore consecutively applied to refine the resulting segments.

Conditional Random Field (CRF) [8] is a kind of statistical modelling method, which has been proved effective in many segmentation tasks - due to its outstanding performance in recovering missing parts and in optimising foreground boundaries [8, 10]. The energy optimisation function we used

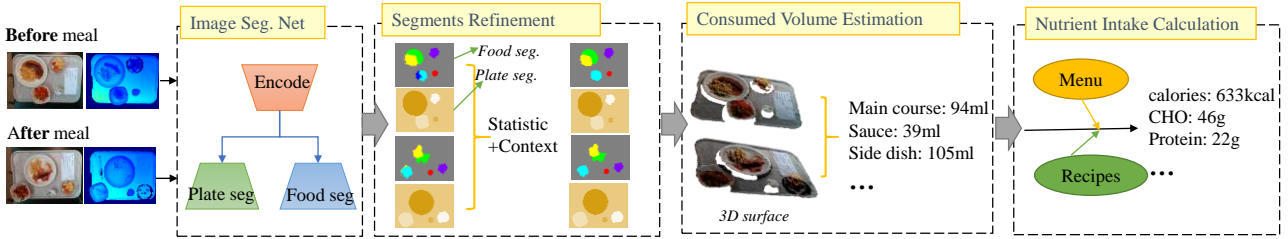

Fig. 3. Overall architecture of the proposed system.

for CRF is expressed by:

$$\min_{\{y_i\}} \sum_i \phi_i(y_i) + \sum_{i,j} \phi_i(y_i, y_j) \quad (1)$$

where $y_i$ indicates food type at the $i^{th}$ pixel, as predicted by the network, $\phi_i(y_i)$ is the unary term of the CRF energy equation that can be expressed as $\phi_i(y_i) = -\log p(y_i)$, where $p(y_i)$ is the probability of $y_i$. $\phi_i(y_i, y_j)$ is the pairwise term that is set to be the standard colour and spatial distance as in [8].

In addition to the use of CRF, the context information provided by the resulted plate segmentation map is also used to further improve the accuracy of food segmentation. This is due to the fact that each food type is usually served with a specific type of plate in the hospital scenario (e.g., the bowl specified for salad in Fig. 2(b)), which is obviously an important clue to improve the food segmentation map. To do so, we utilized a strategy that can be described with the following equation:

$$FoodType = \arg\max_k (\alpha \sum_i p(y_i^k) + \beta \sum_i p(y_i^k \mid yp_i)) \quad (2)$$

In (2), $p(y_i^k)$ indicates the probability of a label belonging to a certain category $k$ at the $i^{th}$ pixel, $yp_i$ is the plate label, and $p(y_i^k|yp_i)$ is the conditional probability of the food label $k$ with plate label $\alpha$ and $\beta$ are hyper parameters.

### C. Estimation of consumed food volume

The consumed volume of each food item is derived by simply subtracting the food volumes before and after the meal, which can be retrieved from the previously available food segmentation map and the depth map, associated with three stages: 1) 3D food surface extraction; 2) plate surface estimation; 3) volume calculation of each food item.

To calculate the volume of each food item, both 3D food surface and the plate surface (i.e. the bottom surface of the food) are required. The depth image is firstly translated into a 3D point cloud and divided into triangular surfaces using the Delaunay triangulation method [9], which can be used to construct the 3D food surface. The plate surface estimation requires the previously built 3D plate model and the plate position, including location and orientation. The location of the plate is estimated using the plate segmentation map, while the plate orientation is set as the normal vector of the tray plane, which is estimated using the RANSAC [11] algorithm. Finally, the volume of each food item is calculated by simply subtracting the food surface and plate surface.

The above method is sufficiently accurate to estimate the consumed volume, provided that the food is served with a normal plate (i.e. those shown in Fig.2 (a)-(d)). However, for packaged containers (shown in Fig. 2 (e)), several additional heuristic methods are required. For instance, although it is difficult to identify the consumption of the sauce (green box labelled) from Fig. 4(a) and (b), applying a heuristic rule, which assumes that sauce consumption is proportional to that of salad, can lead to a reasonable estimation.

### D. Nutrient calculation

From the menu and recipes provided by the Central Kitchen, the detailed food type, weight and the associated nutrient content of each hyper food item is known. The consumed nutrient content of each food item is simply calculated using (3).

$$\text{Consumed nutrient} = \text{total nutrient} \times (1.0 - \frac{\text{volume after meal}}{\text{volume before meal}}) \quad (3)$$

## IV. EXPERIMENTAL RESULT

### A. Experimental setup

As mentioned before, INIMD contains information on 322 meals. In this experiment, we allocate 232 meals for training, 30 meals for validation and the remaining 60 meals for testing. It must be noted that all these 60 meals in the testing set contain as least two image pairs captured before and after the meal.

The segmentation network is trained with the Adam optimiser (lr=0.0002) and categorical cross entropy loss. The initial epoch number is set as 80, and the training process terminates when the validation loss stops decreasing for 10 epochs. The batch size is set as 4. To increase the image variability, we augment RGB-D image pairs by applying left-right and up-down flips during training. Both hyper-parameters ($\alpha$ and $\beta$) in (2) are set as 0.5 in the experiment, while the conditional probability of the food and plate type is calculated based on training set.

The segmentation network module is implemented using Python with Keras (Tensorflow backend), while the other procedures are all implemented using C++.

### B. Food segmentation

The region-based metric [12] is employed to evaluate food segmentation. Both $F_{min}$ and $F_{sum}$ evaluation indicators are used in this work, corresponding to the worst and average segmentation performance, respectively. For both indicators, a higher value means better performance. Here we only evaluate the segmentation performance of the images captured before the meal, since food areas after the meal may be totally disordered and have negligible impact on the food volume estimation, as exemplified in Fig. 4 (red box labelled).

Table II lists the comparison among the segmentation results using different methods. The "MTFCNet (Baseline)" corresponds to the result using the proposed multi-task learning approach, which outperforms that obtained by single-task learning without the "decode plate" component of MTFCNet in Table I, and clearly demonstrates the advantage of using plate information. The additional improvement in performance benefited from post-processing and is also given in the last two rows of Table II, where "+CRF+Context" that employs both CRF and context refinement gives the best result.

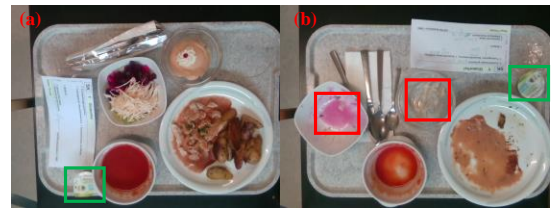

Fig. 4. An example of image pair captured (a) before and (b) after meal consumption

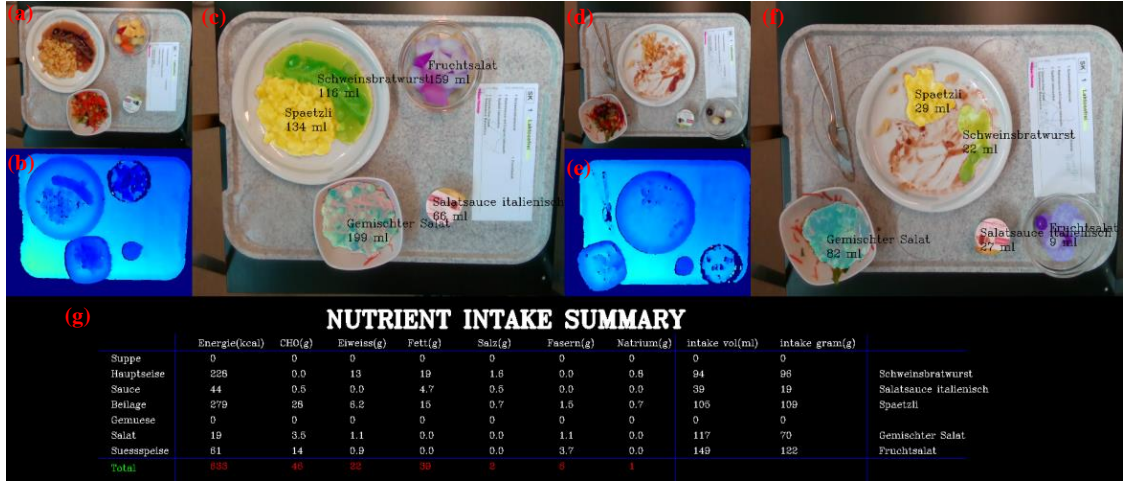

Fig. 5 An example of the proposed system. (a), (b), (d), (e) input colour and depth food tray images before meal and after meal, respectively; (c), (f) food segmentation, volume estimation result before and after meal; (g) output nutrient intake - summary table of the meal.

TABLE II. SEGMENTATION RESULTS

| Method | Fmin (%) | Fsum (%) |
|---|---|---|
| MTFCNet (Baseline) | 61.43 | 83.27 |
| Single-task FCNet | 54.36 | 79.27 |
| +CRF | 66.59 | 86.57 |
| +CRF+Context | **71.59** | **87.04** |

*C. Monitoring nutrient intake*

In this section, the performance of nutrient intake estimation is evaluated on the basis of Mean Absolute Error (MAE) and Mean Relative Error (MRE). Ground truth is calculated by the weight of the plate waste.

The estimated calories, macronutrients, salt, fiber and sodium along with the corresponding errors are reported in Table III. It can be observed that MREs for all items are around 15%, which indicates the contribution of the proposed approach. The final output of the system is illustrated in Fig. 5.

TABLE III. ACCURACY OF NUTRIENT INTAKE ESTIMATION

|  | MAE | MRE (%) |
|---|---|---|
| Calories | 63.78kcal | 12.71 |
| CHO | 6.37g | 12.08 |
| Fat | 3.60g | 13.78 |
| Protein | 2.80g | 17.19 |
| Salt | 0.74g | 15.89 |
| Fiber | 1.06g | 16.87 |
| Sodium | 0.32g | 16.47 |

V. CONCLUSION

In this paper, we have presented the design, development and preliminary evaluation of a novel AI-based system for monitoring nutrient intake in hospitalised patients. Several novel approaches are put forward, such as the new multimedia-nutrient combined database INIMD, the food segmentation technique including the MTFCNet and the post refinement methods. We demonstrated that the system estimates automatically the nutrient intake with a MSE of only 15%.

ACKNOWLEDGEMENT

We would like to thank all the volunteers that contributed in the image acquisition and annotation processes.